\documentclass[12pt]{article}
\usepackage{a4wide}
\usepackage[utf8]{inputenc}
\usepackage[english]{babel}
\usepackage{amssymb,amsmath,mathrsfs}
\usepackage{graphicx}
\usepackage[ruled,vlined]{algorithm2e}
\usepackage{tikz}
\usetikzlibrary{positioning}
\usepackage{hyperref}
\usepackage{url}

\usepackage{theorem}
\newtheorem{definition}{Definition}
\newtheorem{lemma}{Lemma}
\newtheorem{theorem}{Theorem}
\newtheorem{proof}{Proof}

\begin{document}
\title{Optimal spanning tree reconstruction in symbolic regression}
\author{Radoslav G. Neychev, Innokentiy A. Shibaev, and Vadim V. Strijov\footnote{E-mail: vadim@m1p.org}}
\maketitle

\begin{abstract}	
This paper investigates the problem of regression model generation. A model is a superposition of primitive functions. The model structure is described by a weighted colored graph. Each graph vertex corresponds to some primitive function. An edge assigns a superposition of two functions. The weight of an edge equals the probability of superposition. To generate an optimal model one has to reconstruct its structure from its graph adjacency matrix. 
The proposed algorithm reconstructs the~minimum spanning tree from the~weighted colored graph. This paper presents a novel solution based on the prize-collecting Steiner tree algorithm. This algorithm is compared with its alternatives.
\end{abstract}


\section{Introduction}
The symbolic regression is a method to construct a~non-linear model to fit data. The model structure is defined by a superposition of primitive functions. The set of primitive functions is fixed for a particular application problem. Alternative model structures are generated by an optimization algorithm to select an optimal model. This paper proposes to define the model structure by a probabilistic graph. A spanning tree in the graph defines some superposition. To select an optimal model, a minimum spanning tree must be reconstructed from the graph.

The genetic programming methods~\cite{koza1992genetic} %~\cite{davis1991handbook} 
find an optimal subset in the primitive set, but require complex computations. The paper~\cite{searson2010gptips} %,searson2015gptips} 
describe methods that use additional constraints, like linear combinations of the primitive functions. The symbolic regression is a model structure optimization method. Recent achievements are shown in~\cite{stanley2002evolving}.%,wann2019}.

Various methods to solve the the symbolic regression problem are based on the matrix representation of the model structure~\cite{bochkarev2017generation}. However these methods do not include constraints on the number of arguments of the primitive functions and on the graph structure, which delivers admissible superposition.
This paper solves the symbolic regression problem. It requires to reconstruct an admissible superposition from the predicted adjacency matrix with edge probabilities. The~$k$-minimum spanning tree ($k$-MST) reconstruction problem is stated. This problem is NP-hard,
%~\cite{lozovanu1993minimal}, 
so only approximate solutions are applicable~\cite{ravi1996spanning}. The~$k$-MST is equivalent to the prize-collecting Steiner tree (PCST) problem~\cite{chudak2004approximate} due to its equivalence of the relaxed formulation of the linear programming problem statement.
The papers~\cite{ravi1996spanning,awerbuch1998new,arora20062+} present approximate solutions for the~$k$-MST problem. 

\begin{figure}[!hp]
	\centering\small{
   \begin{tikzpicture}[
        roundnode/.style={circle, draw=red!60, fill=red!5, very thick, minimum size=7mm},
        node distance=0.4cm
        ]
        \node[roundnode]            (q_0)                           {$\ast$};
        \node[roundnode]                    (q_1) [below         =of q_0]     {$+$};
        \node[roundnode]                    (q_2) [below left    =of q_1]     {$\ln$};
        \node[roundnode]                    (q_3) [below         =of q_1]     {$x$};
        \node[roundnode]                    (q_4) [below right   =of q_1]     {$\sin$};
        \node[roundnode]                    (q_5) [below         =of q_2]     {$x$};
        \node[roundnode]                    (q_6) [below right   =of q_4]     {$\times$};
        \node[roundnode]                    (q_7) [below left    =of q_6]     {$x$};
        \node[roundnode]                    (q_8) [below         =of q_6]     {$\exp$};
        \node[roundnode]                    (q_9) [below         =of q_8]     {$x$};
        \path[->]
        (q_0)   edge                    node    {}  (q_1)
        (q_1)   edge                    node    {}  (q_2)
        edge                    node    {}  (q_3)
        edge                    node    {}  (q_4)
        (q_2)   edge                    node    {}  (q_5)
        (q_4)   edge                    node    {}  (q_6)
        (q_6)   edge                    node    {}  (q_7)
        edge                    node    {}  (q_8)
        (q_8)   edge                    node    {}  (q_9);
    \end{tikzpicture}}
    \caption{The regression model structure is a directed graph}
    \label{graph}
\end{figure}
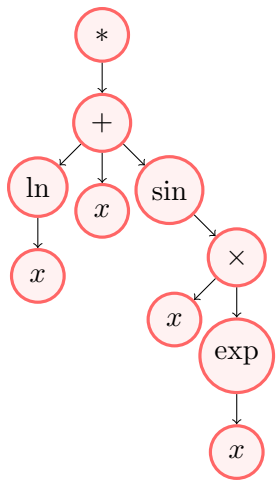

The proposed solution is based on the relaxed version of~$k$-MST problem, which transforms to the PCST problem with constant prizes, the same for all vertices. The fast algorithm for PSCT is described in~\cite{hegde2014fast}. 
%, source code available in~\cite{hegde2015nearly}. 
An alternative solution is based on~$(2-\varepsilon)$ approximation algorithm for PSCT problem. It is compared with the other algorithms, including the tree depth-first traverse, tree breadth-first traverse, Prim's algorithms.

\section{The regression model selection problem}

One has to select a regression model~$\varphi$ from a set of alternative models. The model fits a sample set~$D=\{(x_i,y_i)\}$ and minimises the error 
\begin{equation}
\hat{\varphi}(D)=\mathop{\arg\min}\limits_\varphi\sum_{i=1}^m\bigl(\varphi(x_i)-y_i\bigr)^2.
% \mathbf{f}^{\ast} = \arg\min\limits_{\mathbf{f}\in \mathfrak{F}}L(\mathbf f|\mathbf X,\mathbf y).
\label{task_1}
\end{equation}
The model is a superposition of the primitive functions from some given set. Fig.~\ref{graph} shows an example. 
A structure of the model~$\varphi$, a superposition corresponds to some graph~$G=(V,E)$ where the primitive functions are placed in the vertices~$V$. The {root} vertex is denoted by~$\ast$.  The model is~$\varphi(D) = \ln(x) + x + \sin\bigl(x\times \exp(x)\bigr)$. Its structure as the graph adjacency matrix is provided in Table~\ref{restored_adjacency_matrix}. The primitive functions are listed in the first row. The elements of the matrix are probabilities of the edges~$E$ of the tree. The bold typesetting highlights the edges of the reconstructed tree~$M$, which form the superposition~$\varphi$. To reconstruct the model structure~$\varphi$ as superposition that is defined by the tree~$M$ one needs only the graph representation~$G$ and the primitive functions.

\begin{table}[!htp]
    \centering
        \caption{Probabilities in the adjacency matrix generate the directed graph}
        \begin{tabular}{|c|c|ccccccc|}
            \hline
            arity&prim.&$\ast$&$+$&$\ln$&$\sin$&$\times$&$\exp$&$x$\\
            \hline
            $1$&$\ast$ &0.2&{\bf 0.7}&0.5&0.4&0.5&0.3&0.2\\
            $3$&$+$    &0.3&0.2&{\bf 1.0}&{\bf 0.8}&0.6&0.3&{\bf 0.7}\\
            $1$&$\ln$  &0.3&0.2&0.0&0.0&0.1&0.5&{\bf 0.5}\\
            $1$&$\sin$ &0.1&0.4&0.0&0.5&{\bf 0.9}&0.2&0.5\\
            $2$&$\times$&0.3&0.0&0.3&0.5&0.0&{\bf 0.8}&{\bf 0.6}\\
            $1$&$\exp$ &0.3&0.3&0.4&0.1&0.5&0.4&{\bf 0.4}\\
            \hline
        \end{tabular}
    \label{restored_adjacency_matrix}
\end{table}

State the problem of the model structure reconstruction. There given a collection of sample sets~$\{D_k\}$. Each sample set~$D_k$ corresponds to its model to fit. This model has the structure~$M_k$. So there is a collection of pairs: a sample set and its model structure,~$\{(D_k, M_k)\}$.
Denote by $P$ a map, which predicts the probabilities of nodes in the graph~$G$ using a sample set~$D$. To define a model~$\varphi(D)$ one has to reconstruct the model structure~$M$ from the graph~$G$. Denote by~$R$ this tree reconstruction algorithm. The regression model~$\hat{\varphi}(D)$, which solves the problem~\eqref{task_1}, is defined by 
$
\hat{M}=R\bigl(P(D)\bigr).
$
Since the tree~$M$ plays the main role, this paper sets the quality criterion of the tree reconstruction algorithm as follows: 
\[
\min_{M_k \in G} \frac{1}{K}\sum_{k=1}^K [ \hat{M_k} = M_k)].
\]
The reconstructed tree must be equal to the given tree to guarantee the selected regression model fits its sample set.

\section{The superposition tree reconstruction problem}

The key problem of this work is to propose and analyse the tree reconstruction algorithms. Each reconstructed tree defines a superposition from the previous section. There given the directed weighted graph~$G=(V,E)$ with the colored vertices~$v_i$ and the root vertex~$r$. Every vertex~$v_i \in V$ has its color ~$t(v_i)=t_i$. Every edge~$e_i\in E$ has its weight~\mbox{$w(e_i)=c_i\in[0,1]$}. 

The goal is to reconstruct a minimum-weight directed tree with the root~$r$. It must cover at least~$k$ vertices in the given graph~$G$. And number of tail edges, out-coming from a vertex~$v_i$ of the tree must be less than or equal to~$t_i$. The root~$r$ has one edge,~$t_r=1$.

Formulate this statement in the form of the linear programming problem with integer constraints:
\begin{align}
\underset{\substack{x_e, z_S \\ e\in E, S\subseteq V\backslash \{r\}}}{\text{minimize}}\quad & \sum\limits_{e\in E}c_ex_e \nonumber\\
\text{s.t.\quad} & \sum\limits_{\substack{e\in\delta(S):\\e=(\ast,v_i),~ v_i\in\delta(S)}}x_e + \sum\limits_{T:T\supseteq S}z_T\geqslant 1, & S\subseteq V\backslash \{r\},\nonumber\\
& \sum\limits_{e\in E:~e=(\ast,v)}x_e\leqslant 1, & v\in V,
\label{ilp_our}\\
& \sum\limits_{e\in E:~e=(v,\ast)}x_e\leqslant t_i, & v\in V,\nonumber\\
& \sum\limits_{S\subseteq V\backslash \{r\}}|S|z_S \leqslant n-k,\nonumber\\
& x_e\in\{0,1\}, & e\in E,\nonumber\\
& z_S\in\{0,1\}, & S\subseteq V\backslash \{r\},\nonumber
\end{align}
where~$x_e = 1$ if edge~$e$ is included into the final superposition and~$x_e = 0$ otherwise, $z_S = 1$ for all vertices excluded from the final superposition. Denote by~$e=(\ast, v)$ a~directed edge with the tail~$v$. Denote by~$e=(v, \ast)$  a~directed edge with the head~$v$. 

Every constrain in~\eqref{ilp_our} has its specific purpose.
The first constraint defines the structure of the solution graph as a tree with the root~$r$.
The second constraint defines the orientation of the tree: every vertex has no more than one incoming edge.
The third constraint defines the arity of the used primitive functions, so the number of edges that have a certain vertex as their source is fixed.
The fourth constraint states that the final tree has at least~$k$ vertices. If all weights are non-negative, the fourth constraint on the minimal number of vertices takes a more strict form: the number of vertices should be exactly $k$. However, the weaker constraint allows to find possible connections with other optimization problems. The exact form of the constraints in~\eqref{ilp_our} has the same goal.

\section{The $k\text{-MST}$ and $\text{PCST}$ algorithms for tree reconstruction} 

\begin{definition}{$k$-minimum spanning tree ($k\text{-MST}$).}
There given a weighted graph~$G=(V,E)$ with root~$r$ and edge weights~$w(e_i)=c_i\geqslant 0,~e_i\in E$.
Construct a minimum-weight directed tree with root~$r$, which covers at least~$k$ vertices in~$G$.
\end{definition}
If the same problem is formulated for the directed graphs, the final tree with root~$r$ should be directed. The linear programming problem for the directed~$k\text{-MST}$ excludes the third condition in~\eqref{ilp_our}. % and takes the form:
%%%%%%%\begin{align}
%%%%%%%\underset{\substack{x_e,z_S \\ e\in E, S\subseteq V\backslash \{r\}}}{\text{minimize}}\quad & \sum\limits_{e\in E}c_ex_e \nonumber\\
%%%%%%%\text{s.t.\quad} & \sum\limits_{\substack{e\in\delta(S):\\e=(\ast,v_i),~ v_i\in\delta(S)}}x_e + \sum\limits_{T:T\supseteq S}z_T\geqslant 1, & S\subseteq V\backslash \{r\},\nonumber\\
%%%%%%%& \sum\limits_{e\in E:~e=(\ast,v)}x_e\leqslant 1, & v\in V,
%%%%%%%\label{ilp_k_mst_ord}\\
%%%%%%%& \sum\limits_{S\subseteq V\backslash \{r\}}|S|z_S \leqslant n-k,\nonumber\\
%%%%%%%& x_e\in\{0,1\}, & e\in E,\nonumber\\
%%%%%%%& z_S\in\{0,1\}, & S\subseteq V\backslash \{r\}.\nonumber
%%%%%%%\end{align}
In such form the~$k\text{-MST}$ problem is different from the original superposition tree reconstruction problem~\eqref{ilp_our} by the absence of the third constraint on the arity of primitive functions. It is equivalent to the constraint on number of edges out-coming from a vertex.

\begin{definition}{Prize-collecting Steiner tree ($\text{PCST}$).}
There given a weighted graph~$G=(V,E)$ with root~$r$ and edge weights~$w(e_i)=c_i\geqslant 0,~e_i\in E$, where every vertex~$v_i \in V$ is assigned with a \emph{prize}~$\pi(v_i)=\pi_i\geqslant 0$. Construct a tree~$T$ with root~$r$, which minimizes the  functional: \[\sum\limits_{e\in E}c_ex_e + \sum\limits_{S\subseteq V\backslash\{r\}}\pi(S)z_S,\]
where~$x_e\in\{0, 1\},~x_e=1$ if~$e\in E$ is included in the three~$T$, $z_S\in\{0, 1\},~z_S=1$ for all vertices excluded from tree~$T$ $S = V\backslash V(T)$ and~$\pi(S)= \sum_{v\in S}\pi(v)$.
\end{definition}
In case of directed graphs this problem generalizes to the~asymmetric $\text{A}$-$\text{PCST}$ problem. The linear programming problem for~$\text{A}$-$\text{PCST}$ takes the  form:
\begin{align}
\underset{\substack{x_e,z_S \\ e\in E, S\subseteq V\backslash \{r\}}}{\text{minimize}}\quad & \sum\limits_{e\in E}c_ex_e + \sum\limits_{S\subseteq V\backslash\{r\}}\pi(S)z_S \nonumber\\
\text{s.t.\quad} & \sum\limits_{\substack{e\in\delta(S):\\e=(\ast,v_i),~ v_i\in\delta(S)}}x_e + \sum\limits_{T:T\supseteq S}z_T\geqslant 1, & S\subseteq V\backslash \{r\},\nonumber\\
& \sum\limits_{e\in E:~e=(\ast,v)}x_e\leqslant 1, & v\in V,
\label{ilp_pcst_ord}\\
& x_e\in\{0,1\}, & e\in E,\nonumber\\
& z_S\in\{0,1\}, & S\subseteq V\backslash \{r\}.\nonumber
\end{align}

If the last constraint from%
~\eqref{ilp_our}
%~(\ref{ilp_k_mst_ord}) 
is included into the optimized functional, the $k\text{-MST}$ and~$\text{A}$-$\text{PCST}$ problems have equivalent constraints and only differ in the optimized functional. Such transformation is possible according to the Karush-Kuhn-Tucker conditions and~\cite{ras2017approximate}. If the prize values are equivalent~$\pi(v) = \lambda$ the only difference is the constant term~$\lambda(n-k)$. So the optimization problems~$k$-MST and A-PCST take the forms:
% \begin{align*}
% \underset{\substack{x_e,z_S \\ e\in E, S\subseteq V\backslash \{r\}}}{\text{minimize}}\quad & \sum\limits_{e\in E}c_ex_e + \lambda\left(\sum\limits_{S\subseteq V\backslash \{r\}}|S|z_S - (n-k)\right) & (k\text{-}MST) \nonumber\\
% \underset{\substack{x_e,z_S \\ e\in E, S\subseteq V\backslash \{r\}}}{\text{minimize}}\quad & \sum\limits_{e\in E}c_ex_e + \sum\limits_{S\subseteq V\backslash\{r\}}\pi(S)z_S & (A\text{-}PCST) \nonumber\\
% \end{align*}
\begin{align*}
\underset{\substack{x_e,z_S \\ e\in E, S\subseteq V\backslash \{r\}}}{\text{minimize}}\quad & \sum\limits_{e\in E}c_ex_e + \lambda\left(\sum\limits_{S\subseteq V\backslash \{r\}}|S|z_S - (n-k)\right),\\ % & (k\text{-}MST) \nonumber\\
\underset{\substack{x_e,z_S \\ e\in E, S\subseteq V\backslash \{r\}}}{\text{minimize}}\quad & \sum\limits_{e\in E}c_ex_e + \lambda\sum\limits_{S\subseteq V\backslash\{r\}}|S|z_S. % & (A\text{-}PCST) \nonumber\\
\end{align*}
The constant~$\lambda$ stands for non-negative Lagrange multiplier in the~$k\text{-MST}$ problem and for vertex prize in the~$\text{A}$-$\text{PCST}$. There are several algorithms to solve the~$\text{PCST}$ problem, but not~$\text{A}$-$\text{PCST}$. A possible workaround is to release the constraints on the graph orientation so the~$\text{PCST}$ algorithm could reconstruct the tree orientation later.
% If the constraints on arities are not present, the original problem of reconstructing the superposition tree is reduced to the $k\text{-MST}$ problem on a directed graph. If all prizes are equivalent, this problem is reduced to $\text{A}$-$\text{PCST}$. 

\section{The $(2-\varepsilon)$-approximation algorithm for constrained forest problem}

An overview of techniques for constrained forest problems is provided in~\cite{goemans1995general}. This research selects relevant  results. There given a weighted undirected graph~$G=(V,E)$. All its weights~$w(e_i)=c_i\geqslant 0,~ e_i\in E$. There given some function~$f:2^{V}\to \{0, 1\}$. State the linear programming problem with integer constraints:
\begin{align}
\underset{x_e:~e\in E}{\text{minimize}}\quad & \sum\limits_{e\in E}c_ex_e \nonumber\\
\text{s.t.\quad} & x\bigl(\delta(S)\bigr)\geqslant f(S), & S \subset V, \; S \not= \emptyset,
\label{ilp_cfp}\\
& x_e\in\{0,1\}, & e\in E,\nonumber
\end{align}
where~$x\bigl(\delta(S)\bigr)=\sum\limits_{e\in \delta(S)}x_e$;~$x_e=1$ if edge~$e$ is included into the final set. The function~$\delta(S)$ stands for all edges from~$E$ such that only one of the connected vertices is included in~$S$.

Assume the map~$f$, which satisfies
\[
f(V) = 0, \underbrace{f(S)=f(V\backslash S)}_{\text{symmetry}}, 
\underbrace{A,B\subset V:~A\cap B = \emptyset,~f(A)=f(B)=0\to f(A\cup B) = 0}_{\text{disjunctivity}}.
\]
In these conditions are satisfied, $f$ specifies the number of edges, which starts in the set of vertices~$S$. For example, for the minimum %ideal 
matching problem~$f (S) = 1$ if and only if~\mbox{$|S|\mod 2 = 1$.}

\begin{lemma}
  Let $B\subseteq S\subset V$. Then $f(S) = 0$ and $f(B) = 0$ leads to $f(S\backslash B) = 0$.
  \label{lem1}
\end{lemma}
A problem with such description is the \emph{optimal forest search problem with correct constraints}. Such problem statement \eqref{ilp_cfp} with appropriate map $f$ fits many well-known weighted graph problems, e.g. minimum backbone search, $st$-path, the Steiner problem on the minimum tree. The last problem is NP-complete, so apply an approximate algorithm.

\begin{definition} {($\alpha$-approximation algorithm)}
A heuristic polynomial algorithm that delivers a solution for some optimization problem is called $\alpha$-approximation if it guarantees a constraint-satisfying solution to this optimization problem with a factor less or equal to $\alpha$, so the solution is different from the optimal one no more that by $\alpha$ times in terms of the optimized functional.
\end{definition}

To propose an appropriate approximate algorithm, the integer constraints in~\eqref{ilp_cfp} should be relaxed: 
\begin{align}
\underset{x_e:~e\in E}{\text{minimize}}\quad & \sum\limits_{e\in E}c_ex_e \nonumber\\
\text{s.t.\quad} & \sum\limits_{e\in \delta(S)}x_e\geqslant f(S), & S \subset V, \; S \not= \emptyset,
\label{rlp_cfp}\\
& x_e>0, & e\in E,\nonumber
\end{align}
The dual problem takes the  form:
\begin{align}
\underset{y_S:~S \subset V, \; S \not= \emptyset}{\text{maximize}}\quad & \sum\limits_{S\subset V}f(S)y_S \nonumber\\
\text{s.t.\quad} & \sum\limits_{S:~e\in \delta(S)}y_S\leqslant c_e, & e\in E,
\label{rd_cfp}\\
& y_S>0, &S \subset V, \; S \not= \emptyset,\nonumber
\end{align}
regarding a complementary slackness condition:
%\[
$
y_S\cdot \left(\sum\limits_{e\in \delta(S)}x_e - f(S)\right) = 0, \quad S\subset V.
$%\]

Denote the set of vertices $A=\{v\in V: f(\{v\})=1\}$. Propose an adaptive greedy $\left(2-\frac{2}{|A|}\right)$~---~approximation algorithm for problems of the form~\eqref{ilp_cfp}. The algorithm consists of two stages. On the first stage it greedily combines clusters of vertices increasing the dual variables $y_S$. Initially every vertex belongs to its own cluster. If the next edge $e$ reaches equality in the constraints in~\eqref{rd_cfp}, this edge is added to the set $S$ and the connected clusters will be merged. This stage is similar to Kruskal minimum spanning tree algorithm. %Opposed to the Kruskal algorithm, the effective weight of the next edge is minimized, not the weight itself. This change makes the algorithm capable of adapting to specific data.
In the second stage some edges are removed from the final set $S$. If the edge pruning does not violate the constraints, this edge is to be removed.

The pseudo-code for the described algorithm is provided in the Appendix of this paper. The index $Z_{\text{DRLP}}$ in Algorithm~\ref{alg_general} stands for dual-relaxed linear programming. The initial value of $F:=\emptyset$ in~\ref{alg_general} is equivalent to the assumption $x_e = 0 \;\;\; e \in E$. According to the slackness conditions $y_S = 0,~ S \subset V, \; S \not= \emptyset$.

At any step of the algorithm, cluster $\mathcal{C}$ contains two components $\mathcal{C} = \mathcal{C}_i \cup \mathcal{C}_a$, where $C\in\mathcal{C}_a$ if $f(C) = 1$ and $C\in\mathcal{C}_i$ otherwise. Let's call $\mathcal{C}_a$ an active component.
The variables $d(v)$ in this algorithm are related to the variables $y_S$ from~\eqref{rd_cfp} as~$d(i) = \sum\limits_{S:i\in S}y_S.$ %This statement is proved by induction.

Analyse two different components $C_q,~C_p,~C_q\cap C_p=\emptyset$ on some iteration of the first stage of the algorithm. All $y_S$ should be evenly by some $\varepsilon$ without violating the constraints 
 \[\sum\limits_ {S:~e\in \delta(S)}y_S\leqslant c_e. \] 
In terms of $d(v)$, this condition takes the form~$\sum\limits_{S:~e\in \delta(S)}y_S = d(v_1)+d(v_2),~e=(v_1,v_2),$
so $y_S=0$ for any $S$, such that $v_1, v_2\in S$ because the components only grow on the first stage. Increasing some of the components by $\varepsilon$ leads to the  equation
 \[d(v_1)+d(v_2)+\varepsilon\cdot \bigl(f(C_q)+f(C_p)\bigr)\leqslant c_e,~e=(v_1,v_2), \]
which leads to the formula used in line~$10$ of the Algorithm~\ref{alg_general}. In the case when the next edge is included into the component, the sum $\sum\limits_{S:~e\in \delta (S)}y_S$ will not increase, so the constraints are satisfied.

Edges that can be removed from $F$ without addition of new active components are removed on the second stage of the algorithm. The following lemma defines properties of connected components in $F'$.

\begin{lemma}
  For every connected component $N$ from $F'$ the  equation holds: $f(N)=0$.
  \label{lem2}
\end{lemma}

The following theorem states that the solution derived by the described algorithm is meeting the constraints of the original linear programming problem.

\begin{theorem}
  The edge set $F'$ derived by Algorithm~\ref{alg_general} meets all the constraints of the original problem~\eqref{ilp_cfp}.
  \label{theorem1}
\end{theorem}
% Meeting the constraints leads to the following inequality (because the dual problem is constraint from below) :
% $$Z_{\text{DRLP}} = \sum_{S\subset V}y_S \leqslant Z_{RLP}^{\ast}\leqslant Z_{LP}^{\ast},$$
% where $Z_{LP}^{\ast}$ is the optimal solution of~(\ref{ilp_cfp}).
% The following theorem describes the properties of Algorithm~\ref{alg_general}.

% The following lemma is needed to prove inequality~\eqref{no_inactive_leafs}.
  \begin{lemma}
    Denote graph $H$ where every vertex corresponds to one of the connected components $C\in \mathcal{C}$ on the fixed step of the algorithm. Edge $(v_1,v_2)$ is present if there exists an edge $\hat{e}$ of the original graph included in $F'$: $\hat{e} \in F'$, so the graph $H$ is a forest. There are no leaf vertices within $H$ such that correspond to inactive vertices in the original graph.
  \label{lemma_small}   
  \end{lemma}

\begin{theorem}
  Algorithm~\ref{alg_general} is an $\alpha$-approximate algorithm for problem~\eqref{ilp_cfp} with $\alpha = 2 - \frac{2}{|A|}$ where $A=\{v\in V: f(\{v\})=1\}$.
  \label{theorem2}
\end{theorem} 
Despite this theoretical basis, there is no appropriate function $f$ to state the $\text{PCST}$ problem as referenced in~\eqref{ilp_cfp}. To be applicable is these conditions, the Algorithm~\ref{alg_general} need several modifications.

\section{The upgraded problem statement for $\text{PCST}$}

As in the $\text{A}$-$\text{PCST}$ case, the relaxed form of the linear programming problem $\text{PCST}$ takes form:
\begin{align}
\underset{\substack{x_e,s_v \\ e\in E, v\in V\backslash \{r\}}}{\text{minimize}}\quad & \sum\limits_{e\in E}c_ex_e + \sum\limits_{v\in V\backslash\{r\}}(1-s_v)\pi_v \nonumber\\
\text{s.t.\quad} & \sum\limits_{e\in\delta(S)}x_e\geqslant s_v, & S\subseteq V\backslash \{r\},~v\in S,\label{rlp_pcst_inord}\\
& x_e\geqslant 0, & e\in E,\nonumber\\
& s_v\geqslant 0, & v\in V\backslash \{r\}.\nonumber
\end{align}
This problem statement is different from the original one~\eqref{ilp_pcst_ord}, but it is possible to align the $k\text{-MST}$ problem with it. Indicators $s_v$ show that vertex $v$ is included in the tree.

The dual problem takes form:
\begin{align}
\underset{\substack{y_S:~S\subset V\backslash\{r\}}}{\text{maximize}}\quad & \sum\limits_{S\in V\backslash\{r\}}y_S \nonumber\\
\text{s.t.\quad} & \sum\limits_{S:e\in\delta(S)}y_S\leqslant c_e , & e\in E,\label{rd_pcst_inord}\\
& \sum\limits_{S\subseteq T}y_S\leqslant \sum\limits_{v\in T}\pi_v, & T\subset V\backslash\{r\},\nonumber\\
& y_S\geqslant 0, & S\subset V\backslash\{r\}.\nonumber
\end{align}

Algorithm~\ref{alg_pcst} solves this problem. It is similar to Algorithm~\ref{alg_general}. The dual variables should be updated at an even rate with additional constraints. Then $\varepsilon$ will take the minimum of two values, according to the both groups of constraints. 
%The function $\lambda$  stays for the indicator of the fact that the new component is active, similarly to the $f$ function in the Algorithm~\ref{alg_general}. 
The broader analysis of the approximation properties of the upgraded algorithm is provided in~\cite{goemans1995general}. Algorithm~\ref{alg_pcst} is an $\alpha$-approximate algorithm for $\text{PCST}$ problem with $\alpha = 2 - \frac{2}{n-1}$, where $n$ is the number of vertices in the graph $G$.

\section{Computational experiment}
The main goal of the experiment is to reconstruct the correct superposition tree.  The algorithms used for reconstruction are listed below.

\begin{figure}[!htp]
  \centering{\includegraphics[scale=0.18]{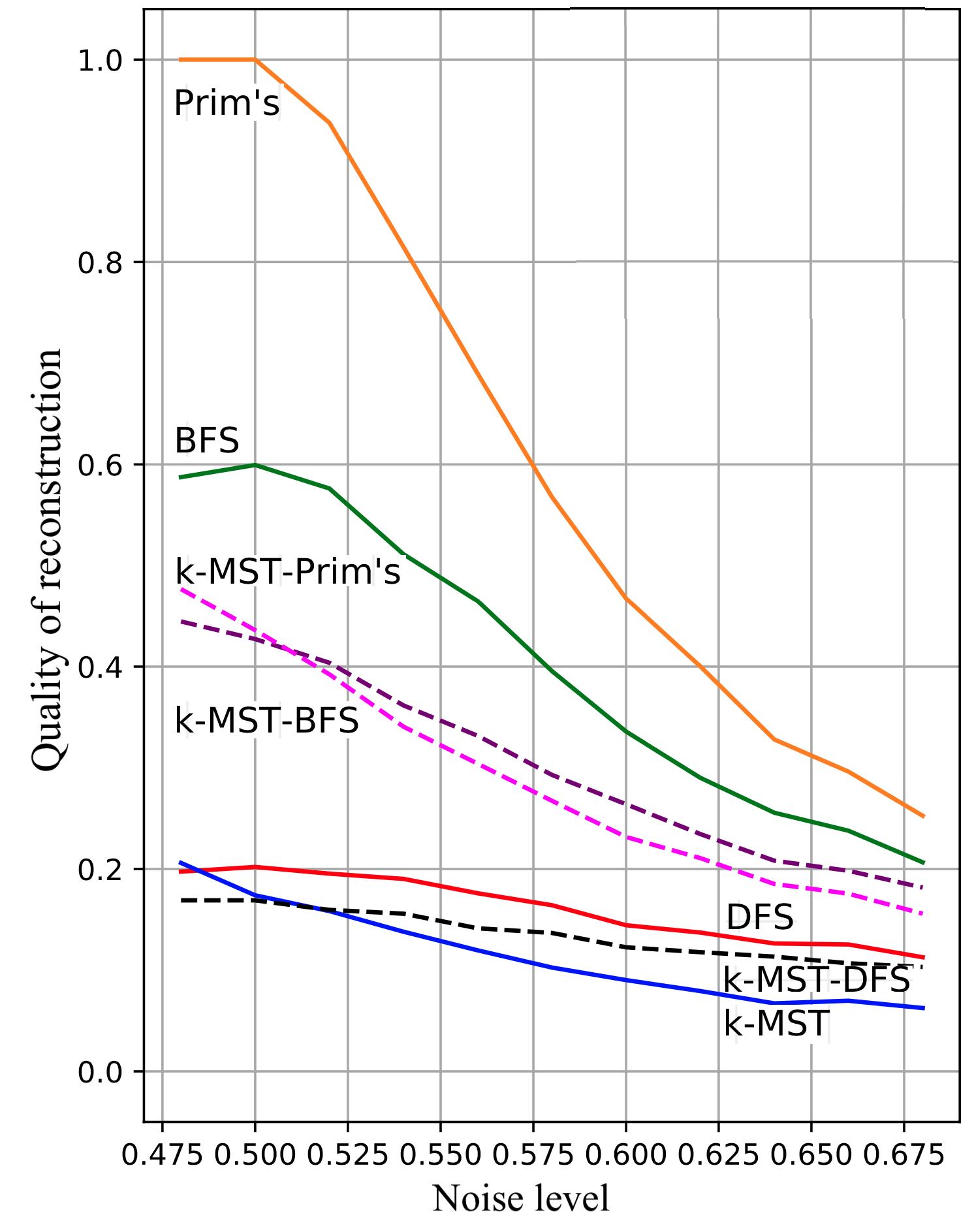}}
  \caption{Quality of the reconstruction algorithms with primitive functions of small arities and unordered inputs}
  \label{ris:main_algs_local}
\end{figure}

\paragraph{DFS, BFS.}
Greedy tree depth-first traverse and Greedy tree breadth-first traverse algorithms. To traverse the edges with highest weights is equivalent to select the most probable path. The traverse algorithm stops when the number of tail edges out-coming from some vertex equals the arity of the corresponding function.

\paragraph{Prim's algorithm.}
This algorithm reconstructs the minimum spanning tree for a graph with additional constraints on the primitive functions' arity. These constraints assign the minimum weight edge. After a vertex is added, all tail edges of this vertex are excluded to preserve direction of the tree. If the number of edges starting in some vertex exceeds the corresponding arity, the rest of edges are excluded from the set of possible edges in this vertex. The algorithm is independent of the traverse procedure. In case of small noise in the adjacency matrix, this algorithm is able to reconstruct the superposition tree without errors. Unfortunately, the algorithm fails if a variable is used several times.

\begin{table}[!htp]
    \caption{Quality of reconstruction algorithms with uniform noise near to $0.5$}
    \centering{
    \begin{tabular}{|l|ccccc|}
      \hline
      $\text{Algorithm, noise}$   &0.50&0.52&0.54&0.56&0.58\\
      \hline
      $\text{DFS}$        &0.20 &0.20 &0.19 &0.18 &0.16\\
      $\text{BFS}$        &0.60 &0.58 &0.51 &0.46 &0.40\\
      $\text{Prim's }$    &1.00 &0.94&0.81&0.69&0.57\\
      $k\text{-MST}$     &0.17 &0.16 &0.14 &0.12 &0.10\\
      $k\text{-MST}$-$\text{DFS}$   &0.17 &0.16 &0.16 &0.14 &0.14 \\
      $k\text{-MST}$-$\text{BFS}$   &0.43 &0.40 &0.36 &0.33 &0.29 \\
      $k\text{-MST}$-Prim's  &0.44 &0.39 &0.34 &0.33 &0.27 \\
      \hline
    \end{tabular}}
    \label{table:main_algs_local}
\end{table}     

\paragraph{Algorithms based on $\text{PCST}$.}
The adjacency matrix~ $M$ must be transformed to the undirected form. Use the square matrix~$M'$ without the last column.  $\text{PCST}$ takes the adjacency matrix~$1 - \frac{1}{2}(M' + M'^{\mathsf{T}})$ with prize value $0.5$ for every vertex. %The subtraction of~$1$ in this case is the element-wise operation. 
The prize value is equal to~$0.5$ because with smaller values the tree will be truncated: if the noise is equal to $0.5$ some vertices might be pruned by error. In case of greater prize values the $\text{PCST}$ tree might include unnecessary vertices. The tree is reconstructed with one of the described algorithms. The results of $\text{PCST}$ can be used as prior for other approaches, $ M':=\frac{1}{2}(M_\text{PCST}' + M')$,
so the $\text{PCST}$ results update~$M'$.

The data generation procedure has the following assumptions: the arities of the function are generated by the Binomial distribution so there are many functions with small arity, all primitive functions have only one input. Any case with partial reconstruction is treated as an error. The quality of reconstruction algorithms is 
%\[
$\frac{1}{K}\sum_{k=1}^K[R\bigl(\bar{N}(M_k)\bigr)=M_k],$%\] 
%\[\text{Acc}\left(R, N, \{M_i\}_{i=1}^N\right) = \frac{1}{N}\sum\limits_{i=1}^N\left[R\bigl(N(M_i)\bigr)=M_i\right],\]
where~$R$ is the reconstruction algorithm and~$\bar{N}=\bigl(N-\min(N)\bigr)/\bigl(\max(N)-\min(N)\bigr)$ is the normalized noise matrix. The matrix~$N$ is generated as~$N(M)=M+U(-\alpha,\alpha)$.
%   $N(M) = \text{Calibration}\bigl( M + U(M.\text{size}, [-\alpha, \alpha])\bigr),$
%  where $\alpha$ as the maximum noise value, 
The random generator returns a matrix with the same shape as $M$ where every element is independent variable from the uniform distribution in the segment~ $[-\alpha,\alpha]$. 
  
Here is the list of seven %$11$ 
compared algorithms: %. The last four algorithms are the cases where $\text{PCST}$ uses a directed graph:
$\text{DFS}$,
$\text{BFS}$,
Prim's algorithm,
$k\text{-MST}$ via $\text{PCST}$,
$k\text{-MST}+\text{DFS}$, 
$k\text{-MST}+\text{BFS}$,
$k\text{-MST}~+$ Prim's algorithm. %,
%$k\text{-MST}$ via $\text{PCST}$ directed,
%$k\text{-MST}+\text{DFS}$ directed,
%$k\text{-MST}+\text{BFS}$ directed,
%$k-MST~+$ Prim's algorithm directed.
Fig.~\ref{ris:main_algs_local} shows the error of the reconstruction algorithms with he noise close to~$0.5$ threshold. The best results are delivered by the Prim's algorithm. The second best solution is based on~$\text{BFS}$. Table~\ref{table:main_algs_local} accompanies Fig.~\ref{ris:main_algs_local} and shows the reconstruction quality of seven algorithms for the border noise values 0.50--0.58.

\section{Conclusion}
This paper proposes and compares different algorithms of superposition reconstruction for the symbolic regression problem. The Prim's algorithm delivers the most accurate results and is the most resistant to small noise in data. 
The proposed algorithm delivers accurate results, but it is more prone to noise in the superposition matrix. The algorithms that are based on BFS and DFS are unable to reconstruct the original superposition with noisy superposition matrices. PCST with BFS used for superposition matrix reconstruction shows baseline results.

%-------------------------------------------------------------------------------------
%-------------------------------------------------------------------------------------
%-------------------------------------------------------------------------------------
%-------------------------------------------------------------------------------------
%-------------------------------------------------------------------------------------
%-------------------------------------------------------------------------------------
\makeatletter
\newcommand{\manuallabel}[2]{\def\@currentlabel{#2}\label{#1}}
\makeatother
\manuallabel{alg_general}{1}
\manuallabel{alg_pcst}{2}
\manuallabel{alg_rec_prim}{3}
\manuallabel{alg_rec_pcst}{4}

%\end{document}
%-----
\clearpage
\newcommand{\lp}{\bigl(}
\newcommand{\rp}{\bigr)}
\section*{Appendix 1: Tree reconstruction algorithms}

\begin{algorithm}[H]
  \DontPrintSemicolon
  \KwData{Weighted undirected graph $G=(V,E)$ with non-negative weights $c_i\geqslant 0$; map $f$}
  \KwResult{Forest $F'$; optimized in problem functional~(\ref{ilp_cfp}) value $Z_{\text{DRLP}}$}
    Stage 1, Merging\;
    \Begin{
        $F \longleftarrow \emptyset$\;
        $Z_{\text{DRLP}} \longleftarrow 0$\;
        $\mathcal{C} \longleftarrow \{\{v\}:v\in V\}$\;
        \ForEach{$v \in V$}{
            $d(v) \longleftarrow 0$\;
        }
        \While{$\exists C\in\mathcal{C}:f(C)=1$}{
            $e^{\ast} = \underset{\substack{e=(i, j):\\ i\in C_p\in\mathcal{C},~j\in C_q\in\mathcal{C},C_p\not = C_q}}{\arg\min}\varepsilon(e)$ where $\varepsilon(e) = \frac{c_e - d(i) - d(j)}{f(C_p)+f(C_q)}$\;
            $F \longleftarrow F\cup {e^{\ast}}$\;
            \ForEach{$C \in \mathcal{C}$}{
                \ForEach{$v \in C$}{
                    $d(v) \longleftarrow d(v) + \varepsilon(e^{\ast})\cdot f(C)$\;
                }
            }
            $Z_{\text{DRLP}} \longleftarrow Z_{\text{DRLP}} + \varepsilon(e^{\ast})\sum_{C\in\mathcal{C}}f(C)$\;
            $\mathcal{C}\longleftarrow \mathcal{C} \backslash \{C_p\}\backslash \{C_q\} \cup \{C_p\cup C_q\}$~~~~~($e^{\ast}$ connects components $C_q$ and $C_p$)\;
        }
    }
    Stage 2, pruning\;
    $F' \longleftarrow \{e\in F:~\exists N\in(V, F\backslash\{e\}),f(N)=1\}$, where $N$ is the connected component\;
    \caption{($2-\varepsilon$)-approximation algorithm for problem~\eqref{ilp_cfp}
    \label{alg_general}}
\end{algorithm}
%-------------------------------------------------------------------------------------
\begin{algorithm}
    \DontPrintSemicolon
    \KwData{Weighted undirected graph $G=(V,E)$ with non-negative edges' weights $c_i\geqslant 0$, prizes $\pi_i\geqslant 0$ and root $r$}
    \KwResult{Tree $F'$ including vertex $r$}
    Stage 1, Merging\;
    \Begin{
        $F \longleftarrow \emptyset$\;
        $Z_{\text{DRLP}} \longleftarrow 0$\;
        $\mathcal{C} \longleftarrow \{\{v\}:v\in V\}$\;
        \ForEach{$v \in V$}{
            Remove markup from $v$\;
            $d(v) \longleftarrow 0$\;
            $w(\{v\}) \longleftarrow 0$\;
            
            \lIf{$v=r$}{$\lambda(\{v\}) \longleftarrow 0$} \lElse{$\lambda(\{v\}) \longleftarrow 1$}
        }
        \While{$\exists C\in\mathcal{C}:\lambda(C)=1$}{
            $e^{\ast} = \underset{\substack{e=(i, j):\\ i\in C_p\in\mathcal{C},~j\in C_q\in\mathcal{C},C_p\not = C_q}}{\arg\min}\varepsilon_1(e)$ where $\varepsilon_1(e) = \frac{c_e - d(i) - d(j)}{\lambda(C_p)+\lambda(C_q)}$\;
            $C^{\ast} = \underset{\substack{C: C\in\mathcal{C},~\lambda(C)=1}}{\arg\min}\varepsilon_2(C)$ where $\varepsilon_2(C) = \sum_{i\in C}\pi_i-w(C)$\;
            $\varepsilon = \min\bigl(\varepsilon_1(e^{\ast}),\varepsilon_2(C^{\ast})\bigr)$\;
            \ForEach{$C \in \mathcal{C}$}{
                $w(C) \longleftarrow w(C) + \varepsilon\cdot \lambda(C)$\;
                \ForEach{$v \in C$}{
                    $d(v) \longleftarrow d(v) + \varepsilon\cdot\lambda(C)$\;
                }
            }
            \uIf{$\varepsilon_1(e^{\ast}) > \varepsilon_2(C^{\ast})$} {
                $\lambda(C^{\ast}) \longleftarrow 0$
                Mark all unmarked vertices from $C^{\ast}$ with $C^{\ast}$.
            } \uElse {
                $F \longleftarrow F\cup {e^{\ast}}$\;
                $\mathcal{C}\longleftarrow \mathcal{C} \backslash \{C_p\}\backslash \{C_q\} \cup \{C_p\cup C_q\}$~~~~~($e^{\ast}$ connects components $C_q$ and $C_p$)\;
                $w(C_p\cup C_q) \longleftarrow w(C_p) + w(C_q)$\;
                \lIf{$r\in C_p\cup C_q$}{$\lambda(C_p\cup C_q) \longleftarrow 0$} \lElse{$\lambda(C_p\cup C_q) \longleftarrow 1$}
            }
        }
    }
    Stage 2, pruning\;
    $F'$ is derived from $F$ by dropping the maximum number of edges meeting the  constraints:
    \begin{enumerate}
        \item All unmarked vertices are connected with root $r$.
        \item If vertex marked with $C$ is connected with root $r$, all other vertices marked with $C$ should be connected with root $r$ as well.
    \end{enumerate}
    \caption{($2-\varepsilon$)-approximate algorithm for $\text{PCST}$ problem\label{alg_pcst}}
\end{algorithm}
%-------------------------------------------------------------------------------------
\begin{algorithm}
    \DontPrintSemicolon
    \KwData{Noised superposition matrix$M\in\mathbb{R}^{n\times (n+1)}_{+}$, list $l$ with $n-1$ arity values for used functions}
    \KwResult{Superposition matrix $M_{res}$ with correct arities}
    \Begin{
        $l \longleftarrow [1] + l$~~~~~(add $1$ to the list)\;
        $M' \longleftarrow $ zero matrix of shape $n\times(n+1)$\;
        $used \longleftarrow \{0\}$\;
        
        $edges \longleftarrow \emptyset$\;
        \ForEach{$j \in range(0,n)$}{
            \If {$j\not\in used$} {
                $edges \longleftarrow edges\cup (0, j, M[0][j])$~~~~~(from, to, weight)\;
            }
        }
        \While{$edges\not = \emptyset$}{
            Find tuple $(from,to,w)$ maximizing the edge weight $w$ over all $edges$\;
            \ForEach{$j \in used$}{
                $M[to][j] = 0$
            }
            \ForEach{$j \in range(0,n)$}{
                \If {$j\not\in used$} {
                    $edges \longleftarrow edges\cup (to, j, M[to][j])$~~~~~(from, to, weight)\;
                }
            }
            \If {$to\not=n$} {
                $edges \longleftarrow edges\backslash (from,to,w)$\;
                $l[from] \longleftarrow l[from] - 1$\;
            }
        
            Remove from $edges$ all tuples $(i,j,w)$ with $j=to$\;
            \If {$l[to]=0$} {
                Remove from $edges$ all tuples $(i,j,w)$ with $i=from$\;
            }
        }
    }
    \caption{Superposition tree reconstruction with Prim's algorithm\label{alg_rec_prim}}
\end{algorithm}
%-------------------------------------------------------------------------------------
\begin{algorithm}
    \DontPrintSemicolon
    \KwData{Noised superposition matrix$M\in\mathbb{R}^{n\times (n+1)}_{+}$, list $l$ with $n-1$ arity values for used functions}
    \KwResult{Superposition matrix $M_{res}$ with correct arities}
    \Begin{
        Drop the last column from matrix $M$ to derive matrix $M'$\;
        $M_{new}' = 1 - \frac{M'+M'^T}{2}$\;
        
        $M_{pcst}' = PCST(M_{new}', 0.5)$\;
        
        Add zero column to the $M_{pcst}'$ on the right to derive $M_{pcst}$\;
        
        Reconstruct the tree from $M_{pcst}$ with some traverse procedure from the root vertex to derive $M_{res}$\;
    }
    \caption{Superposition tree reconstruction with algorithm for the $\text{PCST}$ problem\label{alg_rec_pcst}}
\end{algorithm}
%-------------------------------------------------------------------------------------
%-------------------------------------------------------------------------------------
\clearpage
\section*{Appendix 2: Proofs to the lemmas and theorems}

\begin{proof}[to lemma~\ref{lem1}]
    The Symmetry property leads to $f (V\backslash S) = 0$. Since $V\backslash S \cap B = \emptyset$, the disjunctivity property leads to $f \big((V\backslash S)\cup B\big) = 0$. According to the symmetry property, the  equation holds:  \[f\Big( V\backslash \big((V\backslash S)\cup B\big)\Big) = f(S\backslash B) = 0. \]
\end{proof}
%-------------------------------------------------------------------------------------
\begin{proof}[to lemma~\ref{lem2}]
    Recall that $F'$ is constructed from $F$ via pruning. Hence there is a connected component $C\in F$ such that $N\subseteq C$. The algorithm has stopped, so $f(C) = 0$. All the edges $\delta(N)$ which started from $N$ before pruning and were pruned. Then there is no component $\hat{N}$ such that$f(\hat{N})=1$ present in $(V, E\backslash\{e\}),~e\in\delta(N)$.
    Denote the $C$ components derived via edge pruning from $\delta(N)$ as $N,N_1,\ldots,N_{|\delta(N)|}$), then $f(N_i) = 0$. According to disjunctive property $f\lp\bigcup_{i=1}^{|\delta(S)|} N_i\rp = 0$. Hence, according to Lemma~\ref{lem1}  $f(N) = f\lp V\backslash\bigcup_{i=1}^{|\delta(S)|} N_i\rp = 0$.
\end{proof}
%-------------------------------------------------------------------------------------    
\begin{proof}[to theorem~\ref{theorem1}]
    Assume there is empty set $S \subset V, \; S \not= \emptyset,$ such that $\sum{e\in \delta(S)}x_e < f(S)$.
    Then $f(S) = 1$. For every connected component $C_1,\ldots,C_m$ from $(V,F')$ one of the following equations stand:
    $C_i\subseteq S$ or $C_i\cap S = \emptyset$ according to $\sum_{e\in \delta(S)}x_e=0$. 
    According to Lemma \ref{lem2} and disjunctive property, $f(S) = f\lp\bigcup_{j}C_{i_j}\rp = 0$, which is contrary to the original assumption $f(S) = 1$.
\end{proof}
%-------------------------------------------------------------------------------------
\begin{proof}[to theorem~\ref{theorem2}]
    Recall the  inequality:
     \[
    Z_{LP}^{\ast}\leqslant \sum\limits_{e\in F'}c_e \leqslant \lp 2 - \frac{2}{|A|}\rp Z_{\text{DRLP}}\leqslant 
    \lp 2 - \frac{2}{|A|}\rp Z_{LP}^{\ast}.
     \]
    \begin{enumerate}
        \item The first part holds according to the $Z_{LP}^{\ast}$ definition and the fact that $F'$ meets the constraints according to Theorem \ref{theorem1}.
        \item The last part holds because for the optimal solution $Z_{LP}^{\ast}$ of the problem \ref{ilp_cfp} the following inequality holds due to the below constraints of the dual problem: $Z_{\text{DRLP}} =  \sum_{S\subset V}y_S \leqslant Z_{RLP}^{\ast}\leqslant Z_{LP}^{\ast}$.
        \item The middle part should be proven explicitly.
    \end{enumerate}
    
    After the Algorithm~\ref{alg_general} stops $c_e = \sum y_S$, so the following equation stands:
     \[\sum\limits_{e\in F'}c_e = \sum\limits_{e\in F'}\sum\limits_{S:e\in \delta(S)}y_S = \sum\limits_{S\subset V}y_S\cdot |F'\cap \delta(S)|. \]
    
    So the following inequality should will be proved by induction:
     \[\sum\limits_{e\in F'}c_e = \sum\limits_{S\subset V}y_S\cdot |F'\cap \delta(S)|\leqslant \lp 2 - \frac{2}{|A|}\rp Z_{\text{DRLP}}=\lp 2 - \frac{2}{|A|}\rp \sum\limits_{S\subset V}y_S. \]
    
    \textbf{Basis case:} On the first step of the algorithm $y_S = 0$.
    
    \textbf{Inductive step:} assume the induction hypothesis that for a particular step k, the inequality holds. On the $(k+1)$ step the left hand side is increased by $\varepsilon\sum\limits_{S\in\mathcal{C}_a}|F'\cap \delta(S)|$, where $\mathcal{C}_a$ stays for all active components and  $f(C)=1$. The right hand side will be increased by $\varepsilon\lp 2 - \frac{2}{|A|}\rp\cdot|\mathcal{C}_a|$. Let's focus on the following inequality: 
     \[
    \sum\limits_{S\in\mathcal{C}_a}|F'\cap \delta(S)|\leqslant\lp 2 - \frac{2}{|A|}\rp\cdot|\mathcal{C}_a|.
     \]
    Denote the number of edges starting in $S$ as $d(S) = |F'\cap \delta(S)|$. So
     \[
    \sum\limits_{S\in\mathcal{C}_a}d(S) = \sum\limits_{S\in\mathcal{C}}d(S) - \sum\limits_{S\in\mathcal{C}_i}d(S) \leqslant 2(|\mathcal{C}_a| + |\mathcal{C}_i| -1) - \sum\limits_{S\in\mathcal{C}_i}d(S),
     \]
    where $\mathcal{C} = \mathcal{C}_i\cup \mathcal{C}_a$. The last inequality holds because $F'$ defines a forest in the original graph. The last step is to prove that
    \begin{align}
    \sum\limits_{S\in\mathcal{C}_i}d(S)\geqslant 2|\mathcal{C}_i|,
    \label{no_inactive_leafs}
    \end{align}
    which implies
     \[
    \sum\limits_{S\in\mathcal{C}_a}d(S) \leqslant 2(|\mathcal{C}_a| + |\mathcal{C}_i| -1) - 2|\mathcal{C}_i| = 2\lp 1 - \frac{1}{|\mathcal{C}_a|}\rp\cdot|\mathcal{C}_a|\leqslant2\lp 1 - \frac{1}{|A|}\rp\cdot|\mathcal{C}_a|,
     \]
     which holds because the number of clusters does not increase through time, or equivalently $|A| \geqslant |\mathcal{C}_a|$.
    According to the Lemma \ref{lemma_small} proves the inequality~\eqref{no_inactive_leafs}, so the second part of the original statement is also correct.
\end{proof}     
%-------------------------------------------------------------------------------------    
\begin{proof}[to lemma~\ref{lemma_small}]
		The following lemma is needed to prove inequality~\eqref{no_inactive_leafs}.
         Recall that lone inactive vertices are ignored due to the zero power of every inactive vertex, so they all can be subtracted in the inequality $\sum_{S\in\mathcal{C}}d(S) \leqslant 2(|\mathcal{C}_a| + |\mathcal{C}_i| -1)$.
         
         Assume there is a leaf vertex $v\in V(H)$ connected with edge $e$. This vertex corresponds to the inactive set $C_v\in\mathcal{C}$. This set $C_v$ is included in one of the connected components $N\in F$, where $F$ is the set of vertices before pruning. The fact that $v \in V(H)$ is a leaf implies that all edges connecting $C_v$ with other vertices from $N$ but the edge $e$ were excluded during pruning.
         
         If the edge $e$ is excluded from the component $N$, which is a tree itself, the component splits into $N_1$ and $N_2$. Without loss of generality assume $C_v\subseteq N_1$. The edge $e$ was not pruned, so $f(N_1)=1$ or $f(N_2)=1$. Due to $f(N) = 0$, the only possible variant is $f(N_1) = f(N_2) = 1$. Other cases are contradictory to Lemma~\ref{lem1}.
         
         Denote components $(C_v,C_1,\ldots,C_m)$ derived from $N_1$ if edges from $F'$ are used. For every component but $C_v$ $f(C_i)=0  i \neq v$ because the edges were pruned. But $f(C_v)=0$ according to the original assumption. Hence $f(N_1) = f\lp\bigcup\limits_{i=1}^{m} C_i \cup C_v\rp = 0$ by symmetry, which is contradictory to $f(N_1) = 1$. So there are no leaf vertices in $H$ which correspond to inactive vertices in the original graph, hence power of every inactive vertex is greater or equal to two or equal to zero. In the latter case the vertices do not affect the target inequality.
    \end{proof}
%-------------------------------------------------------------------------------------
\end{document}